\documentclass{ecai2012}

\usepackage[utf8]{inputenc}
\usepackage{macro.EN}
\usepackage{graphicx}
\usepackage{verbatim}

\newcommand{\vsbs}{\vspace{0mm}}
\newcommand{\vsbss}{\vspace{-2mm}}

\begin{document}


\title{Semi-automatic annotation process for procedural texts: An application on cooking recipes}

\author{Valmi Dufour-Lussier\institute{Universit\'e de Lorraine, LORIA, UMR 7503 --- Vand\oe{}uvre-l\`es-Nancy, F-54506, France}
\institute{CNRS, LORIA, UMR 7503 --- Vand\oe{}uvre-l\`es-Nancy, F-54506, France}
\institute{Inria --- Villers-l\`es-Nancy, F-54602, France}
\and Florence Le Ber$^{123}$\institute{Université de Strasbourg/ENGEES, LHYGES, UMR 7517 --- Strasbourg, F-67000, France}
\and \\ Jean Lieber$^{123}$
\and Thomas Meilender$^{123}$\institute{A2ZI --- Commercy, F-55200, France}
\and Emmanuel Nauer$^{123}$
           }

\maketitle
\bibliographystyle{ecai2012}


\begin{abstract}
\taaable is a case-based reasoning system that adapts cooking recipes to user constraints.
Within it, the preparation part of recipes is formalised as a graph.
This graph is a semantic representation of the sequence of instructions composing the cooking process
and is used to compute the procedure adaptation, conjointly with the textual adaptation.
It is composed of cooking actions and ingredients, among others,
represented as vertices, and semantic relations between those,
shown as arcs,
and is built automatically thanks to natural language processing.

The results of the automatic annotation process is often a disconnected graph,
representing an incomplete annotation, or may contain errors. Therefore, a validating and correcting step is required.
In this paper, we present an existing graphic tool named \kcatos, conceived for representing and editing
decision trees, and show how it has been adapted and integrated in \wikitaaable, the semantic 
wiki in which the knowledge used by \taaable is stored. 
This interface provides the wiki users with a way to correct the case representation of the cooking process,
improving at the same time the quality of the knowledge about cooking procedures stored in \wikitaaable.
\end{abstract}

\begin{description}
\item[Keywords:]
cooking,
natural language processing,
procedural texts,
semantic annotation,
semantic wiki.
\end{description}

\vsbs
%
\section{Introduction}
%

\label{sec:introduction}
This paper presents how an automatic textual annotation process of procedural texts, like cooking recipes, can be improved,
using a graphical interface plugin in a wiki and by involving wiki users for correcting and completing the result of the automatic annotation.
This work has been done in the framework of \taaable, a case-based reasoning (\cbr) system which
adapts cooking recipes to user constraints.
According to the user constraints, \taaable looks up, in the recipe base,
whether some recipes satisfy these constraints.
Such recipes, if they exist, are returned to the user; otherwise the system is able to retrieve similar recipes 
(i.e. recipes that match the target query partially),
called source cases in the context of a \cbr application,
and to adapt these recipes, creating new ones.
The knowledge required by \taaable is stored in semantic wiki named \wikitaaable.
Adaptation consists first in substituting some ingredients of the source cases by the ones required by the user.
Then, two additional adaptations are computed: the adaptation of ingredient quantities 
and the adaptation of the text of the preparation procedure~\cite{taaable10}.
This last adaptation requires to represent semantically the sequence of instructions composing the cooking process.
Previous work, based on natural language processing (NLP), has been realised in order to transform 
the textual procedure into a semantic annotation automatically. This semantic annotation takes the form of
a directed graph in which cooking actions and ingredients, among others,
are represented as vertices, and the semantic relations between those are
represented as arcs.

The main objective of this work is to make it possible for wiki users
to edit the graph that is automatically generated for each recipe in \wikitaaable,
in order to correct and eventually complete it. 
In this paper, we show how \kcatos, an existing graphic tool designed to display and edit
decision trees, has been adapted and integrated in \wikitaaable.

The paper is organised as follows:
Section~\ref{sec:Context} specifies the problem in its whole context and introduces \taaable and \wikitaaable.
Section~\ref{sec:TextualCbrAdaptation} introduces the graph representation used for recipes,
the NLP methods used to extract those graphs, and textual adaptation.
Section~\ref{sec:Approach} explains our approach for editing and correcting the graphs using \kcatos.
Section~\ref{sec:Conclusion} concludes the paper.

\vsbs
%
\section{Context and motivation}
\label{sec:Context}
%

\begin{figure}[h]
\begin{center}
\includegraphics[width=\linewidth]{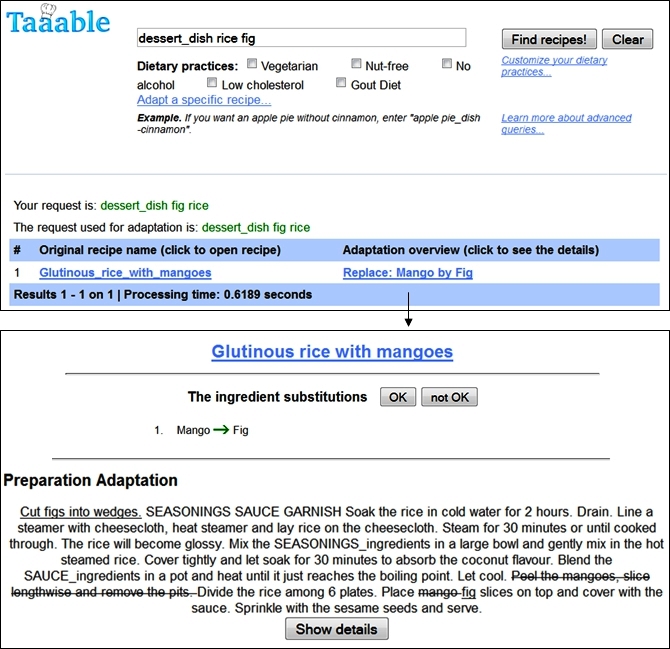}
\vspace{0mm}

\caption{The T{\scriptsize AAABLE} interface. 
Queried for a dessert dish, with rice and fig, T{\scriptsize AAABLE} proposes to replace mango by fig in the
``Glutinous rice with mangoes'' recipe.
After viewing the adapted recipe, the user can give feedback about the substitution 
(``OK'' or ``not OK'').
}
\vspace{-0mm}
\label{Taaable}
\end{center}
\end{figure}

\taaable is a \cbr system that has been designed to take part in the Computer Cooking Contest\footnote{\url{http://computercookingcontest.net/}}, an international contest which aims at comparing 
\cbr system results on a common domain: cooking.
Several challenges are proposed in this contest.
Among them, two challenges (won by \taaable in 2010) are of specific interest to this work:
\vspace{0mm}
\begin{itemize}
\item
the \emph{main challenge}, in which \cbr systems must return recipes satisfying a set of constraints given by the user, 
such as inclusion or rejection of ingredients, the type or the origin of the dish, 
or its compatibility with some diets (vegetarian, nut-free, etc.).
For example: a user may ask for ``a dessert, with rice and fig'', as illustrated in \fig{Taaable}.
Systems have to search into a limited set of (ca. 1500) recipes for recipes satisfying the constraints, 
and if there is no recipe satisfying all the constraints, the systems have to adapt existing recipes into new ones.
\item
the \emph{adaptation challenge}, in which \cbr systems must adapt a given recipe to specific constraints. 
For example, ``adapt the `My strawberry pie' recipe because I do not have strawberries''.
\end{itemize}

\vsbss
\subsection{Principles of {\sc \taaable}}

\vspace{2mm}

Like many \cbr systems~\cite{RiesbeckSchank1989}, \taaable uses an ontology to retrieve 
the source cases that are the most similar to a target case (i.e. the query).
\taaable retrieves and creates cooking recipes by adaptation.
According to the user constraints, the system looks up, in the recipe base (which is a case base),
whether some recipes satisfy these constraints.
Recipes, if they exist, are returned to the user; otherwise the system is able to retrieve similar recipes 
(i.e. recipes that match the target query partially) and adapts these recipes, creating new ones.
Searching similar recipes is guided by several ontologies, i.e. hierarchies of classes 
(ingredient hierarchy, dish type hierarchy, etc.),
in order to relax constraints by generalising the user query.
The goal is to find the most specific generalisation (with the minimal cost) for which recipes exist in the case base.
Adaptation consists in substituting some ingredients of the source cases by the ones required by the user.

\begin{sloppypar}
To deal with the adaptation of a specific recipe (which is the \emph{adaptation challenge} problem), \taaable
uses the same hierarchy based generalisation/specialisation mechanism on a recipe base containing only the recipe
that has to be adapted.
For example, when adapting the ``My Strawberry Pie'' recipe (in which strawberries are required) 
to the constraint ``no strawberry'', $\Strawberry$ is generalised on $\Berry$, which is then specialised in 
another berry (\Raspberry, \Blueberry, \Blackberry, etc.). 
Substitutions (e.g substitute $\Strawberry$ by $\Raspberry$) are proposed to the user.
\end{sloppypar}

\vsbss
\subsection{\sc \wikitaaable}

\wikitaaable\footnote{
	\url{http://wikitaaable.loria.fr}
} is a semantic wiki that uses Semantic MediaWiki~\cite{krotzsch06}  as support for 
encoding knowledge associated to wiki pages.
\wikitaaable contains the set of resources required by the \taaable reasoning system, in  
particular an ontology of the domain of cooking, and recipes.


The cooking ontology is composed of 6 hierarchies: 
a \emph{food} hierarchy (related to ingredients used in recipes, e.g. $\Berry$, $\Meat$, etc.), 
a \emph{dish type} hierarchy (related to the types of recipes, e.g. $\PieDish$, $\Salad$, etc.),
a \emph{dish moment} hierarchy (related to the time for eating a dish, e.g.  $\Snack$, $\Starter$, $\Dessert$, etc.),
a \emph{location} hierarchy (related to the origins of recipes, e.g. $\France$, $\Asia$, etc.),
a \emph{diet} hierarchy (related to food allowed or not for a specific diet, e.g $\Vegetarian$, $\NutFree$, etc.),
an \emph{action} hierarchy (related to cooking actions used for preparing ingredients, $\Cut$, $\Peel$, etc.).
In the semantic wiki, each concept of a hierarchy is encoded as a category page $\texttt{Category:}$ $\texttt{<concept name>}$.
Each concept is described by a short description, lexical variants (used by the annotation bot, for searching concepts
in the full text of recipes), its sub-categories and super-categories.
For berries, the wiki page indicates that $\Berry$ is a sub-concept of $\Fruit$ (corresponding to the $\texttt{Category:Fruit}$ page), 
and sub-categories of $\Berry$ (e.g. $\Raspberry$, $\Blueberry$, etc.) are listed.
Each action has properties, both syntactic and semantic, associated to it, which makes the automatic case acquisition process described hereafter possible.

\begin{figure}[t]
\begin{center}
\includegraphics[width=\linewidth]{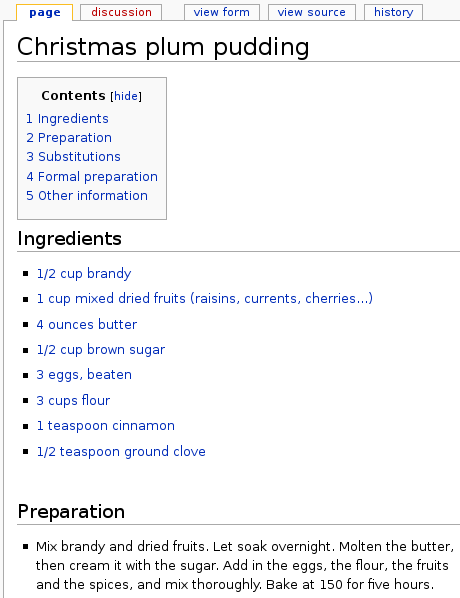}
\vspace{-0mm}
\caption{Example of a W{\scriptsize IKI}T{\scriptsize AAABLE} recipe.}
\vspace{-0mm}
\label{Recipe}
\end{center}
\end{figure}

The set of recipes contained in \wikitaaable are those provided by the Computer Cooking Contest, that have been semantically annotated according to the
domain ontology.  
Each recipe, as the one given in the example of \fig{Recipe}, is encoded as a wiki page, 
composed of several sections: 
a title, which is the name of the recipe, 
an ``\emph{Ingredients}'' section containing the list of ingredients used in the recipe,
each ingredient being linked to its corresponding $Category$ page in the food hierarchy, 
a ``\emph{Textual Preparation}'' section describing the preparation process, 
some possible ``\emph{substitutions}'' which are adaptation knowledge, 
and ``\emph{other information}'' like the dish type, for example.

\begin{figure*}[t]
\begin{center}
\includegraphics[width=\linewidth]{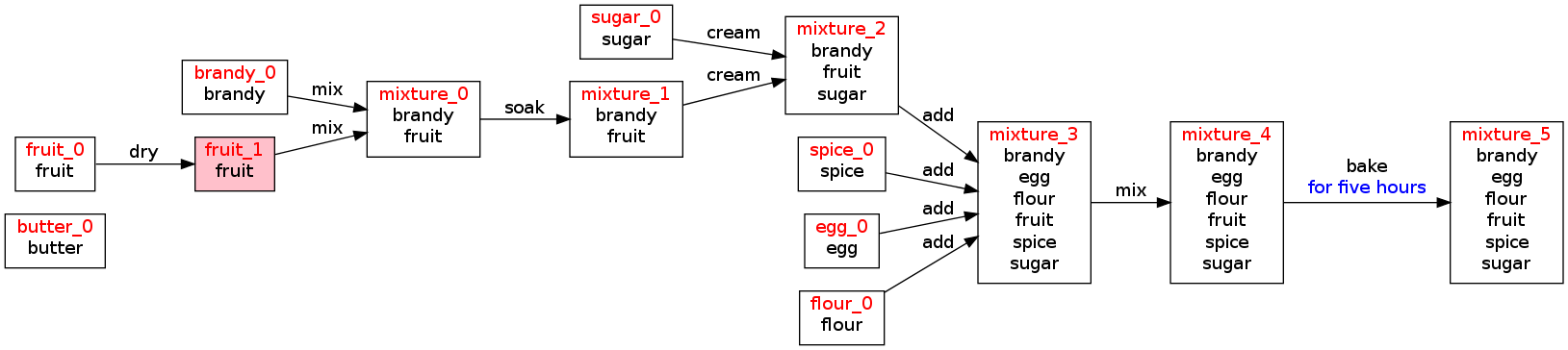}
\vspace{-0mm}
\caption{\label{fig:wt-old}Simple tree representation of the recipe shown in Fig.~\ref{Recipe}.}
\vspace{-0mm}
\end{center}
\end{figure*}

\begin{figure*}[t]
\begin{center}
\includegraphics[width=\linewidth]{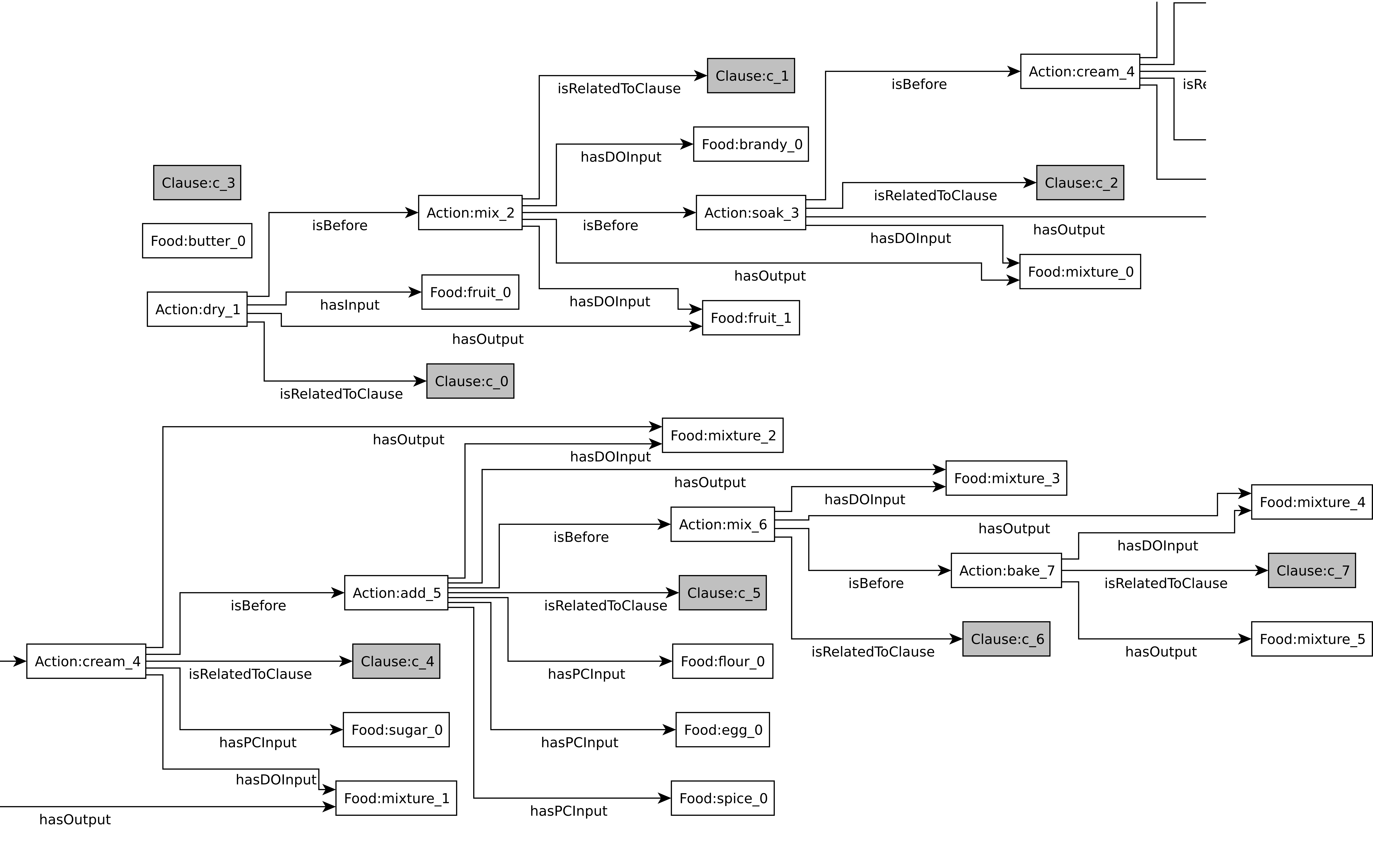}
\vspace{-0mm}
\caption{\label{fig:wt-new}Recipe graph corresponding to the preparation of the recipe given Fig.~\ref{Recipe} in \kcatos (split horizontally for readability).}
\vspace{-0mm}
\end{center}
\end{figure*}

Additionally, for each recipe, there is a graph representing formally the preparation process, which constitutes a highly structured case representation usable by a \cbr engine. For instance, \taaable uses this to propose a fully adapted recipe text to the user. Previously, \wikitaaable used a simple tree representation as shown in Fig.~\ref{fig:wt-old}. This representation is limited and has not been designed to be edited, which is why it is being replaced by the more complete yet easier to edit representation shown in Fig.~\ref{fig:wt-new}.

%

Initially, this graph is acquired from the ingredients and the textual preparation parts through an automatic process making use of NLP methods, which will be presented in the next section.

However, NLP never gives perfect results, and this case acquisition application is no exception. While efforts are being invested in making the acquisition more accurate, if higher quality cases are required in the short term, user intervention is required.

An interface is proposed to make it possible for users to edit the graphs. Graph edition is accomplished in interaction with the automatic acquisition application, such that for each change entered by a user, the system is able automatically to suggest further changes that seem to be required.

Increasing the quantity and quality of the worklfows in the wiki will improve the results of the reasoning system.

The next section details our approach for automatic case acquisition and illustrates how the case representations are being used currently in \taaable.


\vsbs
%
\section{Textual CBR adaptation}
\label{sec:TextualCbrAdaptation}
%

\vsbss
\subsection{Case representation}

Two important methods that are used in \cbr to represent processes, including recipes, are workflows~\cite{minor10iccbr} and qualitative algebras~\cite{dufour12iccbr}. A tree formalism presented earlier was also used in~\cite{dufour10iccbr}. It is important that the representation of a recipe in \wikitaaable be general enough so that it could be translated to any of those formalisms.

To make semi-automatic annotation by users easier, the representation that is used is presented as a graph. However, each recipe graph is directly translatable to an equivalent knowledge base expressed for instance in description logics.
In recipe graphs, each vertex represents the instantiation of a \wikitaaable category, for instance an action or a food item.
An arc expresses a semantic link between two vertices. For one thing, each action vertex is related to a certain amount of food vertices through relations of type \hasDOInput and/or \hasPCInput, and to (generally) one food vertex through the relation \hasOutput. \hasDOInput refers to an input linguistically specified as a direct object, and \hasPCInput as a prepositional complement: e.g. in ``add milk to the batter'', ``milk'' is a direct object and ``batter'' is a prepositional complement. Action vertices are also related to each others through relations such as \isBefore or \isDuring.

Normally, the preparation of a recipe comes to a point where all ingredients are mixed together to form a new whole. This indicates that a correct annotation of a recipe will normally be a connected graph.

\vsbss
\subsection{NLP for procedural case acquisition}

The case acquisition process proposed uses a combination of classical NLP tools which have been slightly adapted to give better results within the framework of procedural texts, as well as some new ideas that were implemented specifically for this type of texts. Other approaches are possible, e.g. \cite{schumacher12www} discusses two different methods based on information extraction to acquire cases as workflows.

A text is first tokenized (split in words), in order to make further linguistic analysis possible. Then, a part-of-speech tag are assigned to each word, indicating whether it is a verb, a noun, etc. This makes is possible to analyse sentences syntax to physically identify, for instance, actions and their arguments. For each action, an action vertex is added to the graph, as well as one (exceptionally, more than one) new food vertex corresponding to its output.

Those steps are not trivial, but effective tools exist, which are either based on stochastic machine learning (e.g. the Brill tagger~\cite{Brill:1992:SRP:1075527.1075553} used for part-of-speech tagging) or rule-based (e.g. the chunker~\cite{abney91chunk} used with hand-crafted grammar rules for syntactic analysis). But this is not sufficient, e.g. to identify properly which food items an action takes as input.

For instance, in a sentence such as ``Peel the mangoes, slice lengthwise and remove the pits'', it is easy as humans to understand that mangoes are being sliced and pitted, but some heuristic is needed is order for a computer to realise that.
This is where the property in the action hierarchy of \wikitaaable comes in handy.
Each action has an arity which makes it possible to tell when an argument is missing. In the sentence above, for instance, ``slice'' requires a direct object which is missing, and ``remove'', a prepositional complement. Whenever this happens, it is taken that the missing argument is in fact the output of the last action, in this example, the mangoes.
In that way, we are able to deal with anaphora, the phenomenon wherein a different word, or no word at all as it might be, is used to represent an object.

Other types of anaphora appear in recipes. The expression ``seasonings ingredients'' clearly refer to some set of food components, so the ingredient hierarchy is used to find all the nodes of the tree that fit under the ``seasonings'' category. A phrase such as ``cover with sauce'' is trickier because there is no obvious clue either in the text or in the ontology as to which food the word ``sauce'' may refer to. We built, from the analysis of thousands of recipes, ``target sets'' of ingredients that usually appear in the food components being referred to by word such as ``sauce'' or, say, ``batter''. This allows for probabilist association of the word with the proper food item with respect to the ingredients of this food.

\vsbss
\subsection{Textual adaptation}

One thing having a high structured case representation of recipes is useful for is to suggest to the user an adaptation at the formal representation level, or even at the text level.

For instance, in~\cite{dufour10iccbr}, an adaptation method that consists in replacing a sequence of actions applicable to an ingredient $\alpha$ with a sequence more suited to a substitution ingredient $\beta$ taken from another recipe was introduced.
Compared with, for instance, complex and error-prone text generation, this method is economical in terms of textual adaptation, because it makes it possible to prune a piece of text and replace it with a piece from a different text, modulo minor linguistic adjustments (fixing capitalisation and punctuation).

Whatever the case representation formalism, the effect of any adaptation method will be to modify the formal representation of the retrieved case to make it suitable as a solution to the target problem. If the solution is to be presented to the user in text format, it is necessary that these modifications be applied to the text at the same time as to its formalised form.

This requires some annotation in the text to establish a mapping between the text and its formal representation. Minimally, clause or sentence segmentation must be marked in the text, so that each clause or sentence can be mapped to the action(s) it expresses. In order simply to replace whatever $\alpha$-specific preparation steps in the recipe with $\beta$-specific preparation steps, this is sufficient.

\begin{figure*}
	\includegraphics[width=\textwidth]{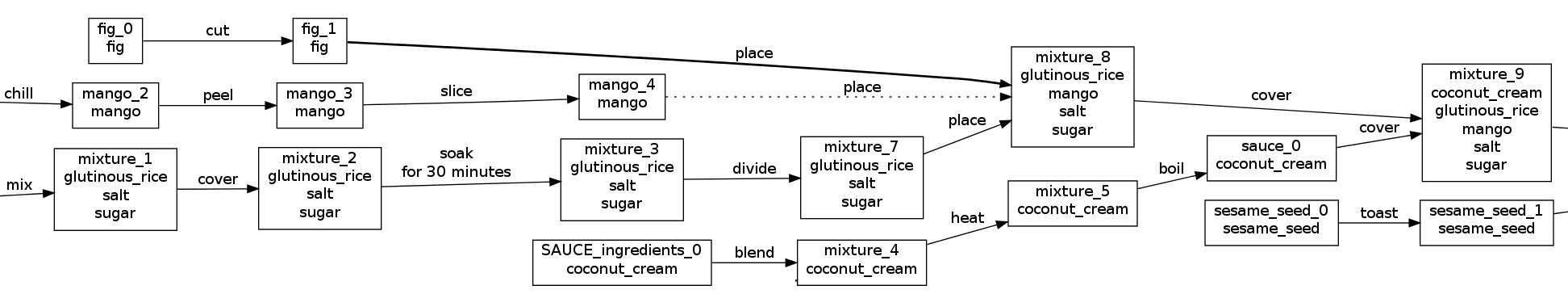}
	\caption{Adaptation by action sequence replacement: a branch related to mango preparation is pruned (dashed arc) and a branch related to fig preparation is grafted (bold arc).}
	\label{fig:graft}
\end{figure*}

Fig.~\ref{fig:graft} shows an example of this adaptation method: the tree branch corresponding to mangoes is being pruned at the arc shown in dotted style, and a fig branch from another recipe is being grated in its place. The corresponding adaptation is done in text at the same time. The result of the textual adaptation appears in Fig.~\ref{Taaable}.

In the new representation format used in \wikitaaable, the text--formal representation mapping is materialised in the graph by one vertex associated each clause from the text. Each action vertex is connected to exactly one clause vertex with an arc labelled with \isRelatedToClause.

\vsbs
%
\section{Editing graphs using \kcatos}
\label{sec:Approach}
%

\vsbss
\subsection{\kcatos}

Many processes or decision makings can be represented using decision trees.
Unfortunately, Semantic Mediawiki does not provide any decision tree editor.
That is why \kcatos has been created.
\kcatos is a semantic decision
tree editor, which provides a collaborative tool to simplify
knowledge acquisition. Using a simple graphical language,
\kcatos allows exporting decision trees to formalised knowledge, 
by proposing an original algorithm export to OWL~\cite{SIMI2012}.

\kcatos is initially a part of a larger work about
collaborative editing of clinical guidelines.
\kcatos has been created as a SMW extension for 
Oncologik\footnote{\url{http://www.oncologik.fr}}~\cite{MeilenderLPJ12}, 
a semantic wiki that shares clinical practice guideline in oncology.
Indeed, to represent decision making, 
guidelines use visual representations that can mostly be viewed 
as decision trees from which a meaning can be extracted. 

\kcatos proposes various features. Among these features, 
a syntactic module can be used to check if the edited tree
respects decision tree rules. Included in the interface,
the module allows to validate trees step by step while drawing,
by identifying shapes with mistakes. As an output, 
different formats are proposed: bitmap (PNG and JPG), vector
graphics (SVG), and ontologies (OWL). Moreover, 
\kcatos includes its own versioning systems. As
each tree is kept on a distant server, modifications are saved. 
Currently, only few functions dealing
with history are available: previous versions of a tree can be viewed and
restored with some information about authors and dates. 
However, some pieces of information are saved into XML
files that will allow to add functionalities such as the comparison of
versions and merging algorithms. Those improvements are
planned to be integrated at a future time.

\kcatos decision tree editor is a web-based application using
Google Web Toolkit\footnote{\url{http://code.google.com/intl/en/webtoolkit/}} (GWT) 
that allows to create complex Ajax applications. 
A few additional APIs dedicated 
to GWT are used to manage the interface. 
Drawing capabilities rely on SVG and JavaScript technologies
while OWL export is done thanks to OWL 
API~\cite{DBLP:journals/semweb/HorridgeB11}. 
Thus, \kcatos is open to collaborative work and web services. Its
framework can be integrated in most of content management
systems.

\subsection{\kcatos for recipes}

\kcatos cannot be used without modifications in \wikitaaable.
One reason is the size of the graphs. \kcatos was made to edit small decision trees, whereas a recipe with $i$ ingredients and $a$ actions will have at least $3a+i$ vertices (each action normally has one for vertex itself, one for its output, and one for its clause).

\begin{figure}
	\includegraphics[width=\columnwidth]{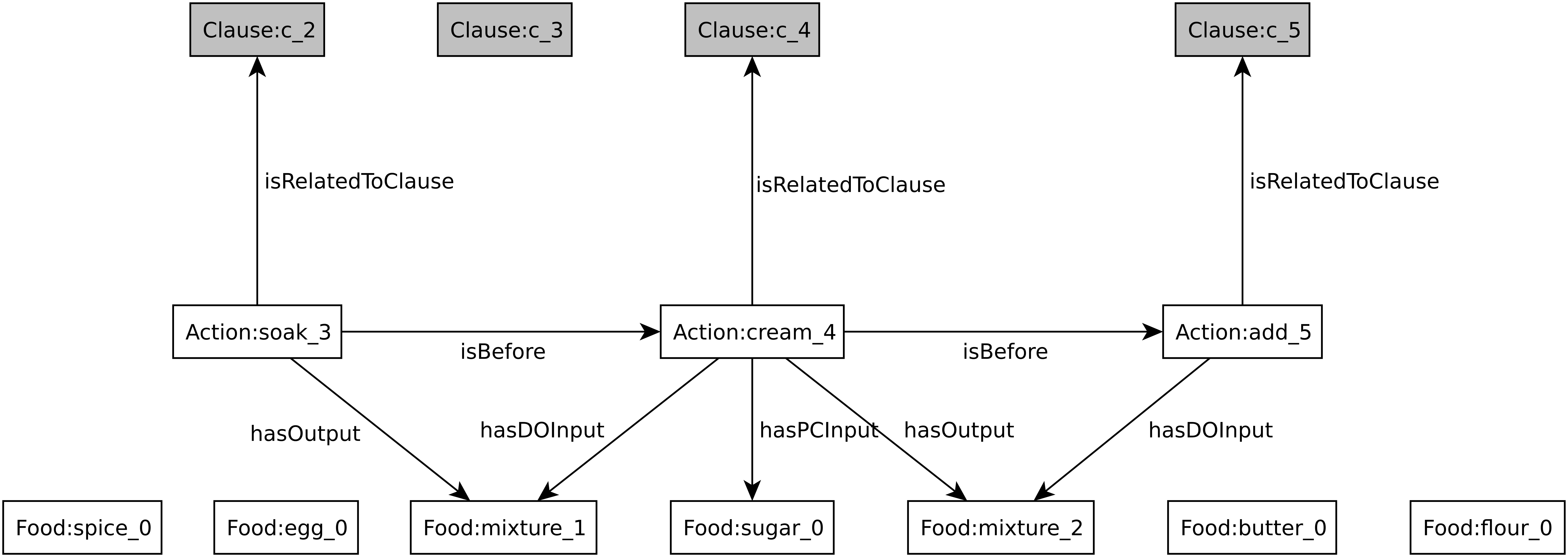}
	\caption{``Intelligent zoom'' on vertex \texttt{Action:cream\_4}.}
	\label{fig:zoom}
\end{figure}

Is is therefore necessary to present the users with features designed to handle the size and complexity of the graphs, such as the ``intelligent zoom'' shown in Fig.~\ref{fig:zoom}. When the user selects an action for which the annotation needs to be corrected, the editor can automatically centre on this action and show only relevant vertices, such as the foods known to be available for use at the time when the action is executed.

Another difficulty is that one simple error in the case acquisition process can result in numerous errors down the road. For instance, considering that ingredient names are not usually repeated more than once in the text, if one input for an action was missed, it will remain missing from the input of all the following actions. Correcting all these mistakes is bound to be a time-consuming and error-prone process.
For this reason, \kcatos was modified to allow for semi-automatic annotation. The user is asked to obey to the text order when correcting mistakes. In exchange, the application is capable of using the human-validated part of the annotation as an additional input and propose a new representation of the recipe, hopefully correcting all the mistakes that were caused by the original error corrected by the user.

\begin{figure}
	\includegraphics[width=\columnwidth]{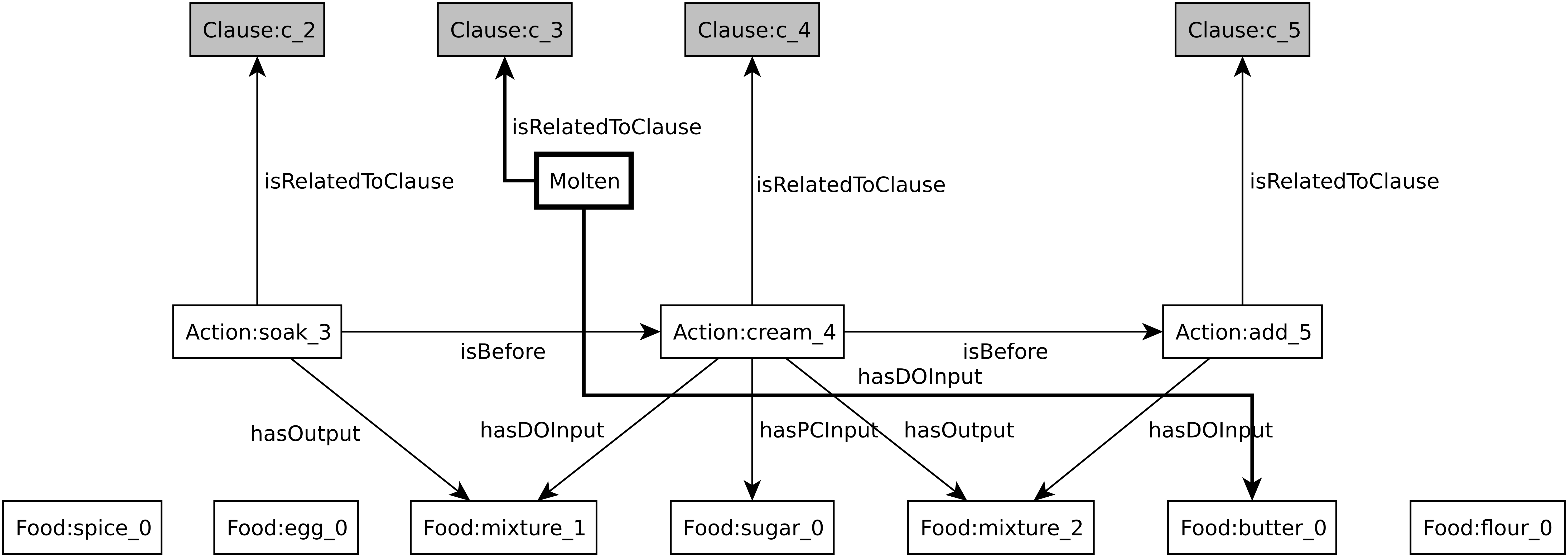}
	\caption{Adding a missing action in \kcatos.}
	\label{fig:correction}
\end{figure}

By looking closely at the graph of Fig.~\ref{fig:wt-new}, it can be seen that the ``molten'' action is missing. This causes the vertices for the ingredient ``butter'' and the clause \texttt{c\_3} (visible in Fig.~\ref{fig:zoom}) to be isolated. This further causes butter to be missing from the input of the action ``cream'', and from the remainder of the recipe. If the user simply adds the missing action and connecting it to the correct input and clause, as shown in Fig.~\ref{fig:correction}, the system is able to correctly repair the rest of the graph.

\vsbs
%
\section{Conclusion}
\label{sec:Conclusion}
%

In order to adapt recipes properly, \taaable uses a highly
 structured case representation for cooking recipes.
This representation is extracted automatically from recipe texts,
 using an application based on natural language processing methods,
 but the results are not perfect.
This paper shows how a graph editing application for semantic wikis,
 \kcatos, can be integrated in \wikitaaable,
 the semantic wiki in which the knowledge required by \taaable is stored,
 in order to provide wiki users with a way to correct the case representation of the cooking process,
 thus improving the quality of the knowledge about cooking procedures stored in \wikitaaable.
Making those tools available to the public will make it possible for us
 to gain feedback about the usability of our proposal and its ability to
 generate correct annotations.

\section*{Acknowledgements}
\kcatos has been conceived and developed in the context of Kasimir project, gathering the company A2ZI, the health network Oncolor, and Université de Lorraine.

\bibliography{biblio}

\end{document}